\documentclass{llncs}

\usepackage{url}
\usepackage{graphicx}
\usepackage{dsfont}
\usepackage{url}
\usepackage{amsfonts}
\usepackage{amsmath}
\usepackage{algorithm}
\usepackage{algpseudocode}	% NOTE: a change was made to icml2018
\usepackage{float}
\usepackage{tabularx,booktabs}
\usepackage[numbers]{natbib}
\usepackage{comment}
\usepackage{ifthen}
\usepackage{hyperref}
\usepackage{mathtools}
\usepackage{rotating}

\newif\iffinal
\finaltrue
\newif\ifarxiv
\arxivtrue

\newcolumntype{Y}{>{\centering\arraybackslash}X}
\newcommand{\env}[1]{\vspace{1mm}\noindent\textbf{#1}}

\newfloat{algorithm}{t}{lop}
\newcommand{\argmax}{\ensuremath{\operatorname{argmax}}}

\newcommand{\eqspace}{\vspace{-3mm}}

\begin{document}

\title{Sample-Efficient Model-Free Reinforcement Learning with Off-Policy Critics}

\author{
	Denis Steckelmacher\inst{1},
	H\'el\`ene Plisnier\inst{1},
	Diederik M. Roijers\inst{2},
	Ann Now\'e\inst{1}
}
\authorrunning{D. Steckelmacher et al.}
\institute{
	Vrije Universiteit Brussel, Brussels, Belgium \and
	Vrije Universiteit Amsterdam, Amsterdam, The Netherlands
	\email{dsteckel@ai.vub.ac.be}
}
\maketitle

\begin{abstract}%
	Value-based reinforcement-learning algorithms provide state-of-the-art results in model-free discrete-action settings, and tend to outperform actor-critic algorithms. We argue that actor-critic algorithms are limited by their need for an \textit{on-policy} critic. We propose Bootstrapped Dual Policy Iteration (BDPI), a novel model-free reinforcement-learning algorithm for continuous states and discrete actions, with an actor and several \textit{off-policy} critics. Off-policy critics are compatible with experience replay, ensuring high sample-efficiency, without the need for off-policy corrections. The actor, by slowly imitating the average greedy policy of the critics, leads to high-quality and state-specific exploration, which we compare to Thompson sampling. Because the actor and critics are fully decoupled, BDPI is remarkably stable, and unusually robust to its hyper-parameters. BDPI is significantly more sample-efficient than Bootstrapped DQN, PPO, and ACKTR, on discrete, continuous and pixel-based tasks. \iffinal Source code: \url{https://github.com/vub-ai-lab/bdpi}. \fi
\end{abstract}

\section{Introduction and Related Work}

State-of-the-art stochastic actor-critic algorithms, used with discrete actions, all share a common trait: the critic $Q^{\pi}$ they learn directly evaluates the actor \citep{Konda1999,Schulman2017,Wu2017,Mnih2016}. Some algorithms allow the agent to execute a policy different from the actor, which the authors refer to as off-policy, but the critic is still on-policy with regards to the actor \citep[for instance]{Haarnoja2018}. ACER and the off-policy actor-critic \citep{Wang2016b,Degris2012} use off-policy corrections to learn $Q^{\pi}$ from past experiences, DDPG learns its critic with an on-policy SARSA-like algorithm \citep{Lillicrap2015}, Q-prop \citep{Gu2017} uses the actor in the critic learning rule to make it on-policy, and PGQL \citep{ODonoghue2017} allows for an off-policy V function, but requires it to be combined with on-policy advantage values. Notable examples of algorithms without an on-policy critic are AlphaGo Zero \citep{Silver2017}, that replaces the critic with a slow-moving target policy learned with tree search, and the Actor-Mimic \citep{Parisotto2016}, that minimizes the cross-entropy between an actor and the Softmax policies of critics (see Section \ref{sec:exp_mimic}). The need of most actor-critic algorithms for an on-policy critic makes them incompatible with state-of-the-art value-based algorithms of the Q-Learning family \citep{Medina2018,Hessel2017}, that are all highly sample-efficient but off-policy. In a \textit{discrete-actions} setting, where off-policy value-based methods can be used, this raises two questions:

\begin{enumerate}
	\vspace{-1mm}
	\item Can we use \textit{off-policy} value-based algorithms in an actor-critic setting?
	\item Would the actor bring anything positive to the agent?
	\vspace{-1mm}
\end{enumerate}

In this paper, we provide a positive answer to these two questions. We introduce Bootstrapped Dual Policy Iteration (BDPI), a novel actor-critic algorithm. Our actor learning rule, inspired by Conservative Policy Iteration (see Sections \ref{sec:background_cpi} and \ref{sec:contrib_actor}), is robust to off-policy critics. Because we lift the requirement for on-policy critics, the full range of value-based methods can now be leveraged by the critic, such as DQN-family algorithms \citep{Hessel2017}, or exploration-focused approaches \citep{Medina2018,Burda2018}. To better isolate the sample-efficiency and exploration properties arising from our actor-critic approach, we use in this paper a simple DQN-family critic. We learn several Q-Functions, as suggested by \cite{Osband2016}, with a novel extension of Q-Learning (see Section \ref{sec:contrib_critic}). Unlike other approaches, that use the critics to compute means and variances \citep{Nikolov2019,Chen2017}, BDPI uses the information in each individual critic to train the actor. We show that our actor learning rule, combined with several off-policy critics, can be compared to bootstrapped Thompson sampling (Section \ref{sec:contrib_thompson}).

Our experimental results in Section \ref{sec:experiments} show that BDPI significantly outperforms state-of-the-art actor-critic \emph{and} critic-only algorithms, such as PPO, ACKTR and Bootstrapped DQN, on a set of discrete, continuous and 3D-rendered tasks. Our ablative study shows that BDPI's actor significantly contributes to its performance and exploration. To the best of our knowledge, this is the first time that, in a discrete-action setting, the benefit of having an actor can be clearly identified. Finally, and perhaps most importantly, BDPI is highly robust to its hyper-parameters, which mitigates the need for endless tuning (see Section \ref{sec:exp_robustness}). BDPI's ease of configuration and sample-efficiency are crucial in many real-world settings, where computing power is not the bottleneck, but data collection is.

\section{Background}

In this section, we introduce and review the various formalisms on which Bootstrapped Dual Policy Iteration builds. We also compare current actor-critic methods with Conservative and Dual Policy Iteration, in Sections \ref{sec:background_pg} and \ref{sec:background_cpi}.

\subsection{Markov Decision Processes}
\label{sec:background_mdp}

A discrete-time Markov Decision Process (MDP) \citep{Bellman1957} with discrete actions is defined by the tuple $\langle S, A, R, T, \gamma \rangle$: a possibly-infinite set $S$ of states; a finite set $A$ of actions; a reward function $R(s_t, a_t, s_{t+1}) \in \mathds{R}$ returning a scalar reward $r_{t+1}$ for each state transition; a transition function $T(s_{t+1} | s_t, a_t) \in [0, 1]$ determining the dynamics of the environment; and the discount factor $0 \leq \gamma < 1$ defining the importance given by the agent to future rewards.

A stochastic stationary policy $\pi(a_t | s_t) \in [0, 1]$ maps each state to a probability distribution over actions. At each time-step, the agent observes $s_t$, selects $a_t \sim \pi(. | s_t)$, then observes $r_{t+1}$ and $s_{t+1}$, which produces an $(s_t, a_t, r_{t+1}, s_{t+1})$ \emph{experience} tuple. An optimal policy $\pi^*$ maximizes the expected cumulative discounted reward $\mathds{E}_{\pi^*}[\sum_t \gamma^t r_t]$. The goal of the agent is to find $\pi^*$ based on its experience within the environment, with no \emph{a-priori} knowledge of $R$ and $T$.

\subsection{Q-Learning, Experience Replay and Clipped DQN}
\label{sec:background_qlearning}

Value-based reinforcement learning algorithms, such as Q-Learning \citep{watkins1992q}, use experience tuples and Equation \ref{eq:qlearning} to learn an action-value function $Q^*$, also called a \emph{critic}, which estimates the expected return for each action in each state when the optimal policy is followed:

\eqspace
\begin{align}
	\label{eq:qlearning}
	Q_{k+1}(s_t, a_t) &= Q_k (s_t, a_t) + \alpha \delta_{k+1} \\
	\nonumber
	\delta_{k+1} &= r_{t+1} + \gamma \max_{a'} Q_k (s_{t+1}, a') - Q_k (s_t, a_t)
\end{align}

\noindent
with $0 < \alpha < 1$ a learning rate. At acting time, the agent selects actions having the largest Q-Value, plus some exploration. To improve sample-efficiency, experience tuples are stored in an \emph{experience buffer}, and are periodically re-sampled for further training using Equation \ref{eq:qlearning} \citep{Lin1992}. Before convergence, Q-Learning tends to over-estimate the Q-Values \citep{Hasselt2010}, as positive errors are propagated by the $\max$ operator of Equation \ref{eq:qlearning}. Clipped DQN \citep{Fujimoto18}, that we use as the basis of our critic learning rule (Section \ref{sec:contrib_critic}), addresses this bias by applying the $\max$ operator to the minimum of the predictions of two independent Q-functions, such that positive errors are removed by the minimum operation. Addressing this over-estimation has been shown to increase sample-efficiency and robustness \citep{Hasselt2010}.

\subsection{Policy Gradient and Actor-Critic Algorithms}
\label{sec:background_pg}

Instead of choosing actions according to Q-Values, Policy Gradient methods \citep{Williams1992,Sutton2000} explicitly learn an \emph{actor} $\pi_\theta(a_t | s_t) \in [0, 1]$, parametrized by a weights vector $\theta$, such as the weights of a neural network. The objective of the agent is to maximize the expected cumulative discounted reward $\mathds{E}_{\pi}[\sum_t \gamma^t r_t]$, which translates to the minimization of Equation \ref{eq:pg} \citep{Sutton2000}:

\eqspace
\begin{align}
	\label{eq:pg}
	\mathcal{L}(\pi_{\theta}) &= -\sum_{t=0}^{T} \mathcal{R}_t \log (\pi_{\theta} (a_t | s_t)) \\
	\label{eq:ac}
	\mathcal{L}(\pi_{\theta}) &= -\sum_{t=0}^{T} Q^{\pi_{\theta}}(s_t, a_t) \log (\pi_{\theta} (a_t | s_t))
\end{align}

\noindent
with $a_t \sim \pi_{\theta}(s_t)$ the action executed at time $t$, and $\mathcal{R}_t = \sum_{\tau=t}^{T} \gamma^{\tau} r_{\tau}$ the Monte-Carlo return from time $t$ onwards. At every training epoch, experiences are used to compute the gradient $\frac{\partial \mathcal{L}}{\partial \theta}$ of Equation \ref{eq:pg}, then the weights of the policy are adjusted by a small step in the opposite direction of the gradient. A second gradient update requires fresh experiences \citep{Sutton2000}, which makes Policy Gradient quite sample-inefficient. Three approaches have been proposed to increase the sample-efficiency of Policy Gradient: trust regions, that allow larger gradient steps to be taken \citep{Schulman2015}, surrogate losses, that prevent divergence if several gradient steps are taken \citep{Schulman2017}, and stochastic\footnote{Deterministic actor-critic methods are slightly different and outside the scope of this paper.} actor-critic methods \citep{Barto1983,Konda1999}, that replace the Monte-Carlo $R_t$ with an estimation of its expectation, $Q^{\pi_{\theta}}(s_t, a_t)$, an \emph{on-policy} critic, shown in Equation \ref{eq:ac}.

The use of $Q^{\pi_{\theta}}$-Values instead of Monte-Carlo returns leads to a gradient of lower variance, and allows actor-critic methods to obtain impressive results on several challenging tasks \citep{Wang2016b,Gruslys2017,Mnih2016}. However, conventional actor-critic algorithms may not provide any benefits over a cleverly-designed critic-only algorithm, see for example \cite{ODonoghue2017}, Section 3.3. Actor-critic algorithms also rely on $Q^{\pi_{\theta}}$ to be accurate for the current actor, even if the actor itself can be distinct from the actual behavior policy of the agent \citep{Degris2012,Wang2016b,Gu2017b}. Failing to ensure this accuracy may cause divergence \citep{Konda1999,Sutton2000}.

\subsection{Conservative and Dual Policy Iteration}
\label{sec:background_cpi}

Approximate Policy Iteration and Dual Policy Iteration are two approaches to Policy Iteration. API repeatedly evaluates a policy $\pi_k$, producing an \emph{on-policy} $Q^{\pi_k}$, then trains $\pi_{k+1}$ to be as close as possible to the greedy policy $\Gamma(Q^{\pi_k})$ \citep{Kakade2002,Scherrer14}. Conservative Policy Iteration (CPI) extends API to \emph{slowly} move $\pi$ towards the greedy policy \citep{Pirotta2013}. Dual Policy Iteration \citep{Sun2018} formalizes as CPI several modern reinforcement learning approaches \citep{Anthony2017,Silver2017}, by replacing the greedy function with a \emph{slow-moving} target policy $\pi'$:

\eqspace
\eqspace
\begin{align}
	&\Gamma(Q^{\pi_k}) \tag{API} \\
	\pi_{k+1} \gets ~ &(1 - \alpha) \pi_k + \alpha \Gamma(Q^{\pi_k}) & \tag{CPI} \\
	&(1 - \alpha) \pi_k + \alpha \pi'_k & \tag{DPI}
\end{align}

\noindent
with $0 < \alpha \le 1$ a learning rate, set to a small value in Conservative Policy Iteration algorithms (0.01 in our experiments). Among CPI algorithms, Safe Policy Iteration \citep{Pirotta2013} dynamically adjusts the learning rate to ensure (with high probability) a monotonic improvement of the policy, while \cite{Thomas2015} propose the use of statistical tests to decide whether to update the policy.

While theoretically promising, CPI algorithms present two important limitations: their convergence is difficult to obtain with function approximation \citep{Wagner2011,BohmerGO16}; and their update rule and associated set of bounds and proofs depend on $Q^{\pi_k}$, an \emph{on-policy} function that would need to be re-computed before every iteration in an on-line setting. As such, CPI algorithms are notoriously difficult to implement, with \cite{Pirotta2013} reporting some of the first empirical results on CPI. Our main contribution, presented in the next section, is inspired by CPI but distinct from it in several key aspects. Our actor learning rule follows the Dual Policy Iteration formalism, with a target policy $\pi'$ built from off-policy critics (see Section \ref{sec:contrib_actor}). The fact that the actor gathers the experiences on which the critics are trained can be compared to the \emph{guidance} that $\pi$ gives to $\pi'$ in the DPI formalism \citep{Sun2018}.

\section{Bootstrapped Dual Policy Iteration}
\label{sec:contrib}

Our main contribution, Bootstrapped Dual Policy Iteration (BDPI), consists of two original components. In Section \ref{sec:contrib_critic}, we introduce an aggressive off-policy critic, inspired by Bootstrapped DQN and Clipped DQN \citep{Osband2016,Fujimoto18}. In Sections \ref{sec:contrib_actor} to \ref{sec:contrib_notcpi}, we introduce an actor that leads to high-quality exploration, further enhancing sample-efficiency. We detail BDPI's exploration properties in Section \ref{sec:contrib_thompson}, before empirically validating our results in a diverse set of environments (Section \ref{sec:experiments}). The complete pseudocode of the algorithm is available in Appendix \ref{sec:app_ppi}, and our implementation of BDPI is available on \url{https://github.com/vub-ai-lab/bdpi}.

\subsection{Aggressive Bootstrapped Clipped DQN}
\label{sec:contrib_critic}

We begin our description of BDPI with the algorithm used to train its critics, Aggressive Bootstrapped Clipped DQN (ABCDQN). Like Bootstrapped DQN \citep{Osband2016}, ABCDQN consists of $N_c > 1$ critics. Combining ABCDQN with an actor is detailed in Section \ref{sec:contrib_actor}. When used without an actor, ABCDQN selects actions by randomly sampling a critic for each episode, then following its greedy function.

Each critic of ABCDQN is trained with an aggressive algorithm loosely inspired by Clipped DQN and Double Q-Learning \citep{Fujimoto18,Hasselt2010}. Each critic maintains two Q-functions, $Q^A$ and $Q^B$. Every \textit{training iteration}, $Q^A$ and $Q^B$ are swapped, then $Q^A$ is trained with Equation \ref{eq:abcdqn} on a set of experiences sampled from an experience buffer, shared by all the critics. Contrary to Clipped DQN, an on-policy algorithm that uses $V(s_{t+1}) \equiv \min_{l=A,B} Q^{l} (s_{t+1}, \pi(s_{t+1}))$ as target value, ABCDQN removes the reference to $\pi(s_{t+1})$ and instead uses the following formulas:

\eqspace
\begin{align}
	\label{eq:abcdqn}
	Q^A_{k+1}(s_t, a_t) &= Q^A_k(s_t, a_t) + \alpha\big( r_{t+1} + \gamma V(s_{t+1}) - Q^A_k(s_t, a_t) \big) \\
	\nonumber
	V(s_{t+1}) &\equiv \min_{l=A,B} Q^{l} \big( s_{t+1}, \argmax_{a'} Q^{A}_k(s_{t+1}, a') \big)
\end{align}

We increase the aggressiveness of ABCDQN by performing several \textit{training iterations} per \textit{training epoch}. Every \textit{training epoch}, every critic is updated using a different batch of experiences, for $N_t > 1$ \textit{training iteration}. As mentioned above, a training iteration consists of applying Equation \ref{eq:abcdqn} on the critic, which produces $Q_{k+1}$ values, either stored in a tabular critic, or used to optimize the parameters of a parametric critic $Q_{\theta}$. The parameters minimize $\sum_{(s, a)} (Q_{\theta}(s, a) - Q_{k+1}(s, a))^2$, using gradient descent for \textit{several} gradient steps.

ABCDQN achieves high sample-efficiency (see Section \ref{sec:experiments}), but its purposefully exaggerated aggressiveness makes it prone to overfitting. We now introduce an actor, that alleviates this problem and leads to high-quality exploration, comparable to Thompson sampling (see Section \ref{sec:contrib_thompson}).

\subsection{Training the Actor with Off-Policy Critics}
\label{sec:contrib_actor}

To improve exploration, and further increase sample-efficiency, we now complement our ABCDQN critic with the second component of BDPI, its actor. The actor $\pi$ takes inspiration from Conservative Policy Iteration \citep{Pirotta2013}, but replaces on-policy estimates of $Q^{\pi}$ with our off-policy ABCDQN critics. Every \textit{training epoch}, after every critic $i$ has been updated on its batch of experiences $E_i \subset B$ uniformly sampled from the experience buffer, the actor is sequentially trained towards the greedy policy of all the critics:

\eqspace
\begin{align}
	\label{eq:ppi_actor}
	\pi(s) &\gets (1 - \lambda) \pi(s) + \lambda \Gamma(Q^{A,i}_{k+1}(s, \cdot)) &\forall~i, \forall~s \in E_i
\end{align}

\noindent
with $\lambda = 1 - e^{-\delta}$ the actor learning rate, computed from the maximum allowed KL-divergence $\delta$ defining a \textit{trust-region} (see Appendix \ref{sec:app_kl}), and $\Gamma$ the greedy function, that returns a policy greedy in $Q^{A,i}$, the $Q^A$ function of the $i$-th critic. Pseudocode for the complete BDPI algorithm is given in Appendix \ref{sec:app_ppi}, and summarized in Algorithm \ref{alg:bdpi}.

\begin{algorithm}[b]
	\caption{Learning with Bootstrapped Dual Policy Iteration (summary)}
	\label{alg:bdpi}
	\begin{algorithmic}
		\For{every critic $i \in [1, N_c]$}
			\State $E \gets$ N experiences sampled from the buffer
			\For{$N_t$ training iterations}
				\State Swap $Q^{A,i}$ and $Q^{B,i}$
				\State Update $Q^{A,i}$ of critic $i$ on $E$ with Equation \ref{eq:abcdqn}
			\EndFor
			\State Update actor on $E$ with Equation \ref{eq:ppi_actor}
		\EndFor
	\end{algorithmic}
\end{algorithm}

Contrary to Conservative Policy Iteration algorithms, and because our critics are off-policy, the greedy function is applied on an estimate of $Q^*$, the optimal Q-function, instead of $Q^{\pi}$. The use of an actor, that slowly imitates approximations of $\Gamma(Q^*) \equiv \pi^*$, leads to an interesting relation between BDPI and Thompson sampling (see Section \ref{sec:contrib_thompson}). While expressed in the tabular form in Equations \ref{eq:abcdqn} and \ref{eq:ppi_actor}, the BDPI update rules produce Q-Values and probability distributions that can directly be used to train any kind of function approximator, on the mean-squared-error loss, and for as many gradient steps as desired. The Policy Distillation literature \citep{Rusu2015} suggests that implementing the actor and critics with neural networks, with the actor having a smaller architecture than the critic, may lead to good results. Large critics reduce bias \citep{Fu2019}, and a small policy has been shown to outperform and generalize better than big policies \citep{Rusu2015}. In this paper, we use actors and critics of the same size, and leave the evaluation of asymmetric architectures to future work.

\subsection{BDPI and standard Conservative Policy Iteration}
\label{sec:contrib_notcpi}

The standard Conservative Policy Iteration update rule (see Section \ref{sec:background_cpi}) updates the actor $\pi$ towards $\Gamma(Q^{\pi})$, the greedy function according to the Q-Values arising from $\pi$. This slow-moving update, and the inter-dependence between $\pi$ and $Q^{\pi}$, allows several properties to be proven \citep{Kakade2002}, and the optimal policy learning rate $\alpha$ to be determined from $Q^{\pi}$ \citep{Pirotta2013}. Because BDPI learns off-policy critics, that can be arbitrarily different from the on-policy $Q^{\pi}$ function, the Approximate Safe Policy Iteration framework \citep{Pirotta2013} would infer an ``optimal'' learning rate of 0. Fortunately, a non-zero learning rate still allows BDPI to learn efficiently. In Section \ref{sec:contrib_thompson}, we show that the off-policy nature of BDPI's critics makes it approximate Thompson sampling, which CPI's on-policy critics do not do. Our experimental results in Section \ref{sec:experiments} further illustrate how BDPI allows fast and robust learning, even in difficult-to-explore environments.

\subsection{BDPI and Thompson Sampling}
\label{sec:contrib_thompson}

In a bandit setting, Thompson sampling \citep{Thompson1933} is regarded as one of the best ways to balance exploration and exploitation \citep{Agrawal2012,Chapelle2011}. Thompson sampling consists of maintaining a posterior belief of how likely any given action is optimal, and drawing actions directly from this probability distribution. In a reinforcement-learning setting, Thompson sampling consists of selecting an action $a$ according to:

\eqspace
\begin{align}
	\label{eq:thompson}
	\pi(a | s) &\equiv P(a = \argmax_{a'} Q^*(s, a'))
\end{align}

\noindent
with $Q^*$ the optimal Q-function. BDPI learns off-policy critics, that produce estimates of $Q^*$. Sampling a critic and updating the actor towards its greedy policy is therefore equivalent to sampling a function $Q \sim P(Q = Q^*)$ \citep{Osband2016}, then updating the actor towards $\Gamma(Q)$, with $\Gamma(Q)(s, a) = \mathds{1}[a = \argmax_{a'} Q(s, a')]$, and $\mathds{1}$ the indicator function. Over several updates, and thanks to a small $\lambda$ learning rate (see Equation \ref{eq:ppi_actor}), the actor learns the expected greedy function of the critics, which (intuitively) folds the indicator function into the sampling of $Q$, leading to an actor that learns $\pi(a | s) = P(a = \argmax_{a'} Q^*(s, a'))$, the Thompson sampling equation for reinforcement learning.

The use of an explicit actor, instead of directly sampling critics and executing actions as Bootstrapped DQN does \citep{Osband2016}, positively impacts BDPI's performance (see Section \ref{sec:experiments}). \cite{Nikolov2019} discuss why Bootstrapped DQN, without an actor, leads to a higher regret than their Information Directed Sampling, and propose to add a Distributional RL \citep{Bellemare2017} component to their agent. \cite{Osband2018} presents arguments against the use of Distributional RL, and instead combines Bootstrapped DQN with prior functions. In the next section, we show that BDPI largely outperforms Boostrapped DQN, along with PPO and ACKTR, without relying on Distributional RL nor prior functions. We believe that having an explicit actor changes the way the posterior is computed, which may positively influence exploration compared to actor-less approaches.

\section{Experiments}
\label{sec:experiments}

To illustrate the properties of BDPI, we compare it to its ablations and a wide range of reinforcement learning algorithms, in four environments with completely different state-spaces and dynamics. Our results demonstrate the high sample-efficiency and exploration quality of BDPI. Moreover, these results are obtained with the same configuration of critics, experience replay and learning rates across environments, which illustrates the ease of configuration of BDPI. In Section \ref{sec:exp_robustness}, we carry out further experiments, that demonstrate that BDPI is more robust to its hyper-parameters than other algorithms. This is key to the application of reinforcement learning to real-world settings, where vast hyper-parameter tuning is often infeasible.

\subsection{Algorithms}
\label{sec:exp_algorithms}

We evaluate the algorithms listed below. We also evaluated ACER and A3C \citep{Wang2016b,Mnih2016}, conventional actor-critic algorithms available in the Open\-AI baselines, but their sample-efficiency was too low for inclusion in our plots.

\begin{center}
\begin{tabular}{lr}
	\textit{BDPI} &  \textit{this paper} \\
	\textit{ABCDQN}, BDPI without an actor \hspace{5mm} & \textit{this paper} \\
	\textit{BDPI w/ AM}, see Section \ref{sec:exp_mimic} & \textit{this paper} \\
	\textit{BDQN}, Bootstrapped DQN & \cite{Osband2016} \\
	\textit{PPO} & \cite{Schulman2017} \\
	\textit{ACKTR} & \cite{Wu2017}
\end{tabular}
\end{center}

Except on \textit{Hallway},\footnote{\url{https://github.com/maximecb/gym-miniworld}} a 3D environment described in the next section, all algorithms use feed-forward neural networks to represent their actor and critic, with one (2 for PPO and ACKTR) hidden layers of 32 neurons (256 on \textit{LunarLander}). The state is one-hot encoded in \textit{FrozenLake}, and directly fed to the network in the other environments. The neural networks are trained with the Adam optimizer \citep{kingma2014adam}, using a learning rate of 0.0001 (0.001 for PPO, ACKTR uses its own optimizer with a varying learning rate). Several extensively-tuned implementations of PPO and ACKTR have been evaluated, to ensure the fairest comparison (parameters in Appendix \ref{sec:app_algos}, we used implementations from \texttt{pytorch-a2c-ppo-acktr} on Github). Unless specified otherwise, BDPI uses $N_c = 16$ critics, all updated every time-step on a different 256-experiences batch, sampled from the same shared experience buffer, for 4 applications of our ABCDQN update rule. BDPI trains its neural networks for 20 epochs per training iteration, on the mean-squared-error loss (even for the policy).

\textit{Hallway} being a 3D environment, the algorithms are configured differently. Changes to BDPI are minimal, as they only consist of using the standard DeepMind convolutional layers, a hidden layer of 256 neurons, and optimizing the networks for 1 epoch per training iteration, instead of 20. PPO and ACKTR, however, see much larger changes. They use the DeepMind layers, 16 replicas of the environment (instead of 1), a learning rate of 0.00005, and perform gradient steps every 80 time-steps (per replica, so 1280 time-steps in total). These PPO and ACKTR parameters are recommended by the author of \textit{Hallway}.

\subsection{BDPI with the Actor-Mimic loss}
\label{sec:exp_mimic}

To the best of our knowledge, the Actor-Mimic \cite{Parisotto2016} is the only actor-critic algorithm, along with BDPI, that learns critics that are off-policy with regards to the actor. We therefore compare BDPI to the Actor-Mimic in Section \ref{sec:exp_results}. These two algorithms perform extremely well, which demonstrates the potential of off-policy critics, with BDPI being more robust than the Actor-Mimic.

The Actor-Mimic is designed for transfer learning tasks. One critic per task is trained, using the off-policy DQN algorithm. Then, the cross-entropy between the actor and the Softmax policies $S(Q_i)$ of all the critics is minimized, using the (simplified) loss of Equation \ref{eq:mimic}.

\eqspace
\begin{align}
	\label{eq:mimic}
	\mathcal{L}(\pi_{\theta}) &= -\sum\limits_{\mathclap{s \in S, a \in A, i < N}} S(Q_i)(a|s) \log (\pi_{\theta} (a | s))
\end{align}

Applying the Actor-Mimic to a single-task setting is possible. We implemented an agent based on BDPI, that retains its ABCDQN critics, but replaces our actor learning rule of Equation \ref{eq:ppi_actor} with the Actor-Mimic loss of Equation \ref{eq:mimic}. Because we only change how the actor is trained, and still use our aggressive critics, we ensure the fairest comparison between our actor learning rule and the cross-entropy loss of the Actor-Mimic. In our experiments, the Actor-Mimic loss with Softmax policies fails to learn efficiently, even after extensive hyper-parameter tuning, probably because the Softmax prevents the policy from becoming deterministic in states where this is necessary. We therefore replaced the Softmax with the greedy function, which led to the much better results that we present in Section \ref{sec:exp_results}.

\subsection{Environments}
\label{sec:exp_environments}

\begin{figure}[t]
	\centering
	\includegraphics{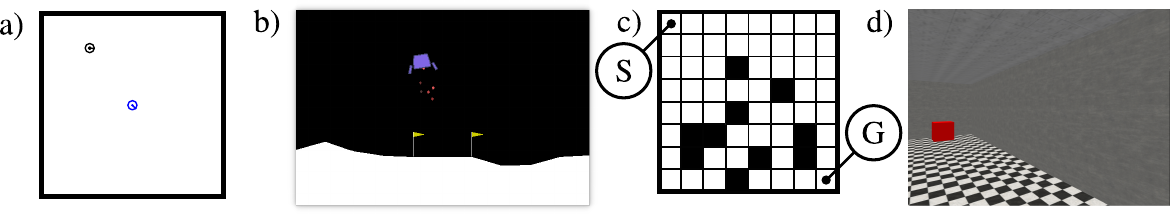}
	\caption{The four environments. a) \textit{Table}, a large continuous-state environment with a black circular robot and a blue charging station. b) \textit{LunarLander}, a continuous-state task based on the Box2D physics simulator. c) \textit{Frozen Lake}, an 8-by-8 slippery gridworld where black squares represent fatal pits. d) \textit{Hallway}, a 3D pixel-based navigation task.}
	\label{fig:envs}
\end{figure}

Our evaluation of BDPI takes place in four environments that challenge the algorithms on different aspects of reinforcement learning: exploration with sparse rewards (\textit{Table}), high-dimensional state-spaces (vector \textit{LunarLander}, pixel-based \textit{Hallway}), and high stochasticity (\textit{FrozenLake}).

\env{Table} simulates a tiny robot on a large table that has to locate its charging station and dock (see Figure \ref{fig:envs}a). The table is a 1-by-1 square. The goal is located at $(0.5, 0.5)$, and the robot always starts at $(0.1, 0.1)$, facing away from the goal. A fixed initial position makes exploration more challenging, as the robot never spawns close to the goal. The robot observes its current $(x, y, \theta)$ position and orientation, with $\theta \in [-\pi, \pi]$. Three actions allow the robot to either move forward 0.005 units, or turn left/right 0.1 radians. A reward of 0 is given every time-step. The episode finishes with a reward of -50 if the robot falls off the table, 0 after 200 time-steps, and 100 when the robot successfully docks, that is, its location is $(0.5 \pm 0.05, 0.5 \pm 0.05, \frac{\pi}{4} \pm 0.3)$. The slow speed of the robot and reward sparsity make \textit{Table} more difficult to explore than most Gym tasks \citep{Gym}.

\env{LunarLander} is a high-dimensional continuous-state physics-based simulation of a rocket landing on the moon (see Figure \ref{fig:envs}b). The agent observes the location and velocities of various components of the lander, and has access to four actions: doing nothing, firing the left/right side engines for one time-step, and firing the main engine. The reward signal for this task is quite complicated but informative, as it directly maps the distance between the rocket and the landing pad to a reward, on every time-step. The environment is considered solved when a cumulative reward of 200 or more is achieved per episode \citep{Gym}.

\env{FrozenLake} is a $8 \times 8$ grid composed of slippery cells, holes, and one goal cell (see Figure \ref{fig:envs}c). The agent can move in four directions (up, down, left or right), with a probability of $\frac{2}{3}$ of actually performing an action other than intended. The agent starts at the top-left corner of the environment, and has to reach the goal at its bottom-right corner. The episode terminates when the agent reaches the goal, resulting in a reward of +1, or falls into a hole, resulting in no reward.

\env{Hallway} is a 3D pixel-based environment, that simulates a camera-based robotic task in the real world. \textit{Hallway} consists of a rectangular room with a target red box, and the agent. The size of the room, location of the goal and initial position of the agent are randomly chosen for each episode. Four discrete actions allow the agent to move forward/backward and turn left/right. Movement is slow, and the amount of movement is stochastic for each time-step. The reward signal is sparse: 0 every time-step, and 1 when the goal is reached. The episode ends with a reward of 0 after 500 time-steps. This sparse reward function heavily stresses the ability of a reinforcement-learning algorithm to train deep convolutional neural networks on small amounts of reward data.

\begin{figure}[t]
	\centering
	\includegraphics[width=\textwidth]{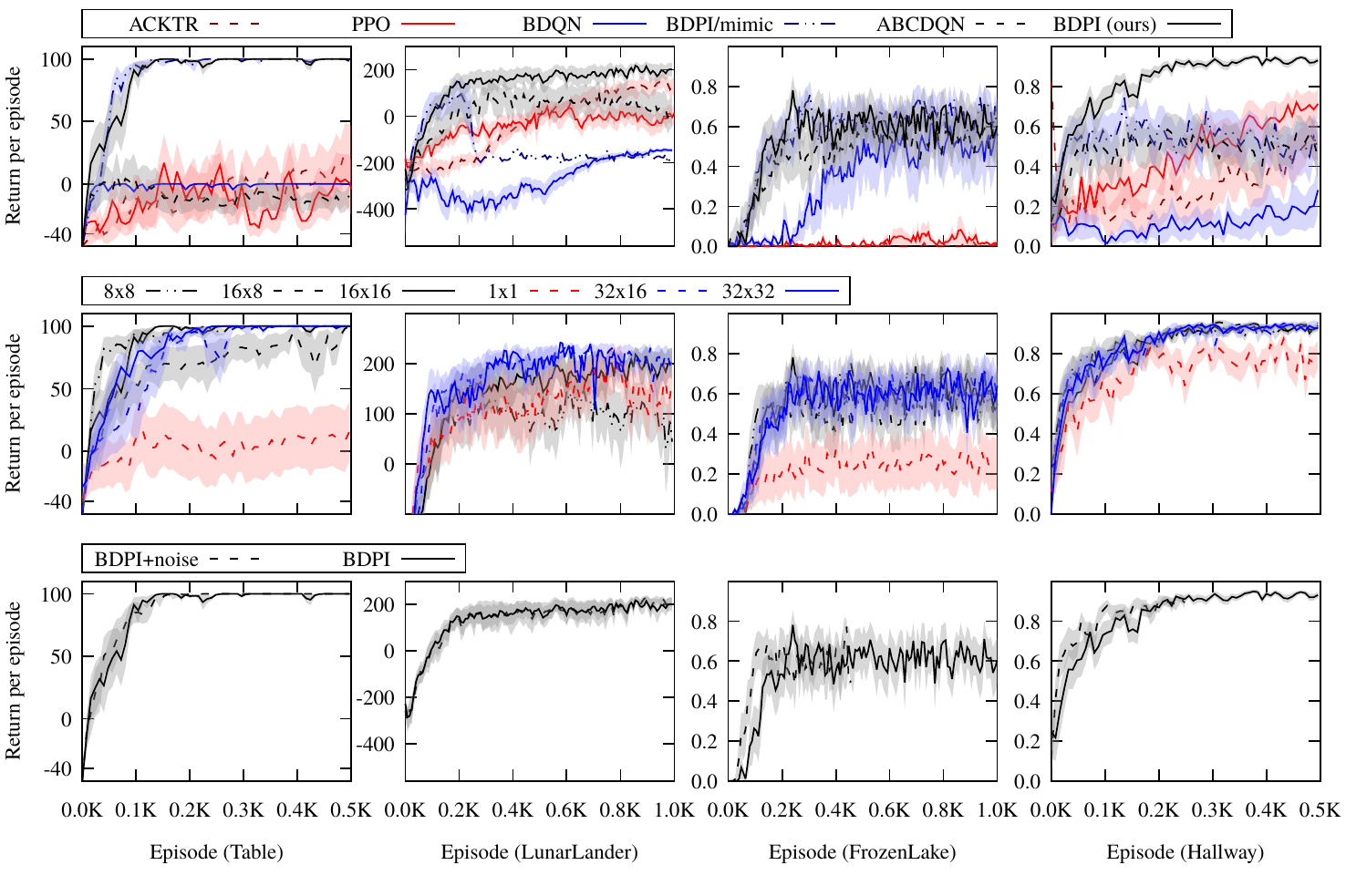}
	\caption{Results on \textit{Table}, \textit{LunarLander}, \textit{FrozenLake} and \textit{Hallway}. \textit{Top:}~BDPI (16 critics, updated for 4 iterations per time-step) outperforms all the other algorithms in every environment, sometimes significantly (\textit{Table} and 3D pixel-based \textit{Hallway}). \textit{Middle:}~Varying the number of critics and how often they are trained, as long as there are more than one critic, only has minimal impact on BDPI's performance, which demonstrates its robustness. \textit{Bottom:}~Adding off-policy noise (see text) does not impact BDPI on any of the environments.}
	\label{fig:results}
\end{figure}

\subsection{Results}
\label{sec:exp_results}

Figure \ref{fig:results} shows the cumulative reward per episode obtained by various agents in our four environments. These results are averaged across 8 runs per agent, with the shaded regions representing the standard error. The plots compare BDPI to the algorithms detailed in Section \ref{sec:exp_algorithms}, and display the effect of varying key hyper-parameters of BDPI.

\env{Algorithms} BDPI is the most sample-efficient of all the algorithms, and also achieves the highest returns (especially on hard-to-explore \textit{Table} and pixel-based \textit{Hallway}). BDPI with the Actor-Mimic loss matches BDPI with our actor learning rule on \textit{Table}, but fails to learn \textit{LunarLander} and \textit{Hallway}. ABCDQN (BDPI without its actor) fails on \textit{Table}, an environment where exploration is key, and is generally inferior to BDPI. These results show that both having an explicit actor, and training it with our update rule of Section \ref{sec:contrib_actor}, are necessary to achieve top performance. Bootstrapped DQN is highly sample-efficient on \textit{FrozenLake}, but does not explore well enough on the other environments. PPO and ACKTR, after extensive tuning and with several implementations tested, are not as sample-efficient as BDPI and Bootstrapped DQN, two off-policy algorithms using experience replay. Even with per-environment hyper-parameters, PPO and ACKTR need about 5K episodes to learn \textit{FrozenLake}, and 1K episodes on \textit{Table}. BDPI is the only algorithm that, with a single configuration for all the environments, automatically adjusts to the complexity of a task to achieve maximum sample-efficiency.

Interestingly, PPO and ACKTR do perform well on 3D \textit{Hallway}. We tentatively point out that, due to the prevalence of pixel-based environments in the modern reinforcement-learning literature, current algorithms and hyper-parameters may focus more on the representation learning problem than on the reinforcement learning aspect of tasks. Also note that on \textit{Hallway}, PPO and ACKTR use 16 replicas of the environment (instead of 1 for BDPI, and PPO/ACKTR on the other environments). This setting greatly stabilizes the algorithms, but cannot be applied to real-world physical robots.

\env{Critics} Increasing the number of critics leads to smoother learning curves in every environment, at the cost of sample-efficiency in \textit{Table}, where a higher variance in the bootstrap distribution of critics seems to help with exploration. Having only one critic seriously degrades BDPI's performance, and having less than 16 critics is detrimental on \textit{LunarLander}, where the environment dynamics are complex. This indicates that more critics are beneficial in complex environments, but may slightly reduce pure exploration.

\env{Off-Policy noise} BDPI's actor learning equations do not refer to any behavior policy or on-policy return, and its critics are learned with a variant of Q-Learning. This hints at BDPI being an off-policy algorithm. We now empirically confirm this intuition. In this experiment, training episodes have, at each time-step, a probability of 0.2 that the agents executes a random action, instead of what the actor wants (0.05 on \textit{Table}, where docking requires precise moves). Testing episodes do not have this noise. The agent learns only from training episodes. Such off-policy noise does not negatively impact BDPI's learning performance. Robustness to off-policy execution is an important property of BDPI for safety-critical tasks with backup policies.

The performance of BDPI, obtained with a single set of hyper-parameters for all the environments\footnote{Only the number of hidden neurons changes between some environments, a trivial change.}, demonstrate BDPI's sample-efficiency, high-quality exploration, and strong robustness to hyper-parameters, as rigorously detailed in the next section.

% Mention that aggressiveness improves results but not that much (improves them quite a bit on Table, not on the other environments)

\subsection{Robustness to Hyper-Parameters}
\label{sec:exp_robustness}

Hyper-parameters often need to be tweaked depending on the environment. Therefore, it is highly desirable that an algorithm provides good performance even if not optimally configured, as BDPI does. To objectively measure an algorithm's robustness to its hyper-parameters, we draw inspiration from sensitivity analysis. Thousands of runs of the algorithm are performed on randomly-sampled configurations of hyper-parameters, with each configuration evaluated on the total reward obtained over 800 episodes on \textit{LunarLander}. Then, we compute the average absolute difference of total reward between random pairs of configurations, weighted by their distance in configuration space. This measures how much changing hyper-parameters affects performance. See Appendix \ref{sec:app_robustness} for more details, and the list of hyper-parameters we consider for each algorithm.

We evaluated numerous algorithms available in the OpenAI baselines. The algorithms, sorted by ascending sensitivity, are DQN with Prioritized ER (930), BDPI (1167), vanilla DQN (1326), A2C (2369), PPO (2452), then ACKTR (5815). Our plot in Appendix \ref{sec:app_robustness} shows that the apparent robustness of DQN-family algorithms comes from them performing equally badly for every configuration. 35\% of BDPI's configurations outperform the best configuration among all the other algorithms.

\section{Conclusion and Future Work}

In this paper, we propose Bootstrapped Dual Policy Iteration (BDPI), an algorithm where a bootstrap distribution of aggressively-trained off-policy critics provides an imitation target for an actor. Multiple critics, combined with our actor learning rule, lead to high-quality exploration, comparable to bootstrapped Thompson sampling. Off-policy critics can be learned with any state-of-the-art value-based algorithm, depending on the application domain. BDPI is easy to implement, and remarkably robust to its hyper-parameters. The hyper-parameters we used for the highly-stochastic \textit{FrozenLake} gridworld allowed BDPI to largely outperform the state of the art on three other environments, one of which pixel-based. This, and the availability of BDPI's full source code, makes it one of the first plug-and-play reinforcement-learning algorithm that can easily be applied to new tasks.

\begin{figure}[b]
	\centering
	\includegraphics{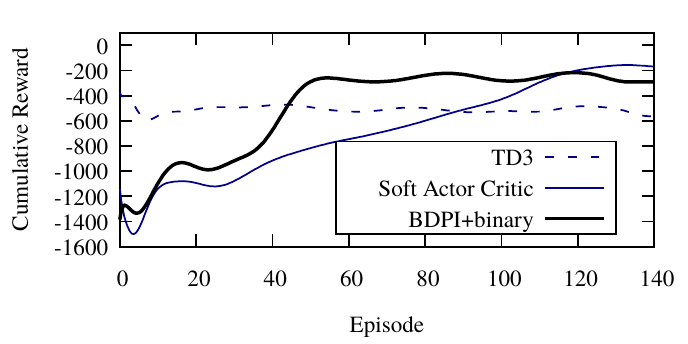}
	\vspace{-3mm}
	\caption{BDPI adjusted for continuous actions with Binary Action Search \citep{Pazis2009} is more sample-efficient than TD3 \cite[seems to quickly learn to spin]{Fujimoto18} and the Soft Actor-Critic \citep{Haarnoja2018} on the Inverted Pendulum task.}
	\label{fig:pendulum}
\end{figure}

While we focus on discrete actions in this paper, the high-quality exploration and robustness to sparse rewards of BDPI lead to encouraging results with discretized continuous action spaces. In Figure \ref{fig:pendulum}, we show that Binary Action Search, an approach that allows precise control of continuous actions, at the cost of increased sparsity in the reward function \citep{Pazis2009}, allows BDPI to outperform the Soft Actor-Critic and TD3, three state-of-the-art continuous-actions algorithms. In future work, we will explore and evaluate various discretization approaches, pursuing the goal of applying BDPI to today's complicated continuous-action tasks.

\iffinal
\section*{Acknowledgments}

The first and second authors are funded by the Science Foundation of Flanders (FWO, Belgium), respectively as 1129319N Aspirant, and 1SA6619N Applied Researcher.
\fi

\setlength{\bibsep}{0pt plus 0.3ex}
\bibliographystyle{splncs04}
\bibliography{biblio}

% This appendix will not be used, it's just here so that cross-references work
% Disable citations in the appendix, to unclobber the citation list. The citations will be taken (in the appendix) from the EWRL paper.
%\renewcommand{\citep}[1]{}
%\renewcommand{\cite}[1]{}
\vfill
\pagebreak
\appendix
\onecolumn
\normalsize

\section{Bootstrapped Dual Policy Iteration Pseudocode}
\label{sec:app_ppi}

The following pseudocode provides a complete description of the BDPI algorithm. To keep our notations simple and general, the pseudocode is given for the tabular setting, and does not refer to any parameter for the actor and critics. An implementation of BDPI based on function approximation, such as the neural networks we use in our experiments, uses the equations below to produce batches of state-action or state-value pairs. The function approximator is then trained on these batches, minimizing the mean-squared-error loss, for several gradient steps.

\begin{algorithm}[h]
	\caption{Bootstrapped Dual Policy Iteration}
	\begin{algorithmic}
		\Require A policy $\pi$
		\Require $N_c$ critics. $Q^{A,i}$ and $Q^{B,i}$ are the two Clipped DQN networks of critic $i$.
		\Procedure{BDPI}{}
			\For{$t \in [1, T]$}
				\State \Call{Act}{}
				\If{$t$ a multiple of $K$}
					\State \Call{Learn}{}
				\EndIf
			\EndFor
		\EndProcedure
		\Procedure{Act}{}
			\State Observe $s_t$
			\State Draw $a_t \sim \pi(s_t)$
			\State Execute $a_t$, observe $r_{t+1}$ and $s_{t+1}$
			\State Add $(s_t, a_t, r_{t+1}, s_{t+1})$ to the experience buffer
		\EndProcedure
		\Procedure{Learn}{}
			\For{every critic $i \in [1, N_c]$ (in random order)} \Comment{Bootstrapped DQN}
				\State Sample a batch $E$ of $N$ experiences from the experience buffer
				\For{$N_t$ iterations} \Comment{Aggressive BDQN}
					\ForAll{$(s_t, a_t, r_{t+1}, s_{t+1}) \in E$} \Comment{Clipped DQN}
						\State $\hat{Q}(s_t, a_t) \gets r_{t+1} + \gamma \min_{l=A,B} Q^{l,i} (s_{t+1}, \argmax_{a'} Q^{A,i}(s_{t+1}, a'))$
					\EndFor
					\State Train $Q^{A,i}$ towards $\hat{Q}$ with learning rate $\alpha$
					\State Swap $Q^{A,i}$ and $Q^{B,i}$
				\EndFor
				\State $\pi \gets (1 - \lambda) \pi + \lambda \Gamma(Q^{A,i})$ \Comment{CPI with an off-policy critic}
			\EndFor
		\EndProcedure
	\end{algorithmic}
\end{algorithm}

\section{The CPI Learning Rate Implements a Trust-Region}
\label{sec:app_kl}

A trust-region, successfully used in a reinforcement-learning algorithm by \cite{Schulman2015}, is a constrain on the Kullback-Leibler divergence between a policy $\pi_k$ and an updated policy $\pi_{k+1}$. In BDPI, we want to find a policy learning rate $\lambda$ such that $D_{KL}(\pi_k || \pi_{k+1}) \le \delta$, with $\delta$ the \emph{trust-region}.

While a trust-region is expressed in terms of the \emph{KL-divergence}, Conservative Policy Iteration algorithms, the family of algorithms to which BDPI belongs, naturally implement a bound on the \emph{total variation} between $\pi$ and $\pi_{k+1}$:

\begin{align}
	\nonumber
	\pi_{k+1} &= (1 - \lambda) \pi_k + \lambda \pi' &\text{see Equation \ref{eq:ppi_actor} in the paper} \\
	\label{eq:tv}
	D_{TV}(\pi_{k+1}(s) || \pi_k(s)) &= \sum_a |\pi_{k+1}(a | s) - \pi_k(a | s)| \\
	\nonumber
	&\le 2\lambda
\end{align}

The total variation is maximum when $\pi'$, the target policy, and $\pi_k$, both have an action selected with a probability of $1$, and the action is not the same. In CPI algorithms, the target policy is a greedy policy, that selects one action with a probability of one. The condition can therefore be slightly simplified: the total variation is maximized if $\pi_k$ assigns a probability of 1 to an action that is not the greedy one. In this case, the total variation is $2\lambda$ (2 elements of the sum of (\ref{eq:tv}) are equal to $\lambda$).

The Pinsker inequality \citep{Pinsker1960} provides a lower bound on the KL-divergence based on the total variation. The inverse problem, upper-bounding the KL-divergence based on the total variation, is known as the Reverse Pinsker Inequality. It allows to implement a trust-region, as $D_{KL} \le f(D_{TV})$ and $D_{TV} \le 2\lambda$, with $f(D_{TV})$ a function applied to the total variation so that the reverse Pinsker inequality holds. Upper-bounding the KL-divergence to some $\delta$ then amounts to upper-bounding $f(D_{TV}) \le \delta$, which translates to $\lambda \le \frac{1}{2} f^{-1}(\delta)$.

The main problem is finding $f^{-1}$. The reverse Pinsker inequality is still an open problem, with increasingly tighter but complicated bounds being proposed \citep{Sason2015}. A tight bound is important to allow a large learning rate, but the currently-proposed bounds are almost impossible to inverse in a way that produces a tractable $f^{-1}$ function. We therefore propose our own bound, designed specifically for a CPI algorithm, slightly less tight than state-of-the-art bounds, but trivial to inverse.

If we consider two actions, we can produce a policy $\pi_k(s) = \{0, 1\}$ and a greedy target policy $\pi'(s) = \{1, 0\}$. The updated policy $\pi_{k+1} = (1 - \lambda) \pi_k + \lambda \pi'$ is, for state $s$, $\pi_{k+1}(s) = \{\lambda, 1 - \lambda\}$. The KL-divergence between $\pi_k$ and $\pi_{k+1}$ is:

\begin{align}
	\nonumber
	D_{KL}(\pi_k || \pi_{k+1}) &= 1 \log \frac{1}{1 - \lambda} + 0 \log \frac{0}{\lambda} \\
	\label{eq:bound}
	&= \log \frac{1}{1 - \lambda}
\end{align}

\noindent
if we assume that $\lim_{x \rightarrow 0} x \log x = 0$. Based on the reverse Pinsker inequality, we assume that if the two policies used above are greedy in different actions, and therefore have a maximal total variation, then their KL-divergence is also maximal. We use this result to introduce a trust region:

\begin{align*}
	D_{KL}(\pi_k || \pi_{k+1}) &\le \delta & \text{trust region} \\
	\log \frac{1}{1 - \lambda} &\le \delta \\
	\frac{1}{1 - \lambda} &\le e^{\delta} \\
	\lambda &\le 1 - e^{-\delta}
\end{align*}

\noindent
Interestingly, for small values of $\delta$, as they should be in a practical implementation of BDPI, $1 - e^{-\delta} \approx \delta$. The trust-region is therefore implemented by choosing $\lambda = \delta$, which is much simpler than the line-search method proposed by \cite{Schulman2015}.

\subsection{State-Dependent Exploration}
\label{sec:app_entropy}

\begin{figure}
	\centering
	\includegraphics[]{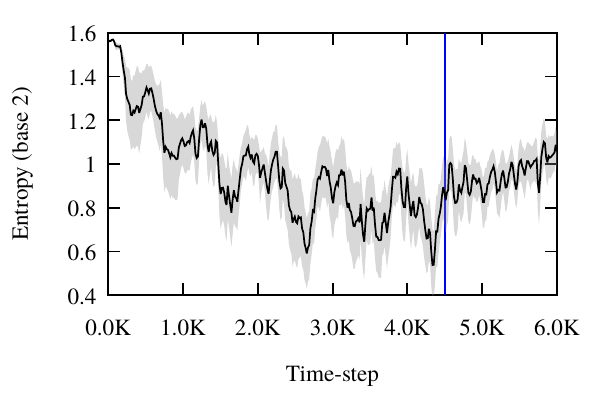}
	\caption{Entropy of the policy per time-step, on the \textit{Table} environment (running average and standard deviation of 8 runs). The entropy oscillates as the agent falls off the table, which resets the environment to familiar states. After some time (blue bar), runs start learning distinct policies, whose entropies cannot be observed anymore on an averaged plot.}
	\label{fig:entropy}
\end{figure}

Compared to Bootstrapped DQN, well-known for its high-quality exploration, BDPI lacks an important component: explicit deep exploration. Deep exploration consists of performing a sequence of directed exploration steps, instead of exploring in a random direction at each time-step \citep{Osband2016}. Bootstrapped DQN achieves deep exploration by greedily following a single critic, sampled at random, for an entire \textit{episode}. BDPI trains its actor towards a randomly-selected critic at every \textit{time-step}, which is incompatible with deep exploration. We empirically show in Section \ref{sec:exp_results} that BDPI outperforms Bootstrapped DQN, so the loss of explicit deep exploration does not seem to negatively affect performance. In Figure \ref{fig:entropy}, we provide a likely explanation in the \textit{Table} environment. At the early stages of training, the agent regularly falls off the table, which resets the episode. This can be observed as dips in the entropy of the actor. We believe that this is caused by a sort of novelty-based exploration, probably more limited than what highly-advanced algorithms produce \citep{Burda2018}, but still present. After a few episodes, the individual runs learn different policies, which breaks the correlation between them and explains the flat portion of Figure \ref{fig:entropy}. The emergence of such an interesting exploration strategy, leading to higher-quality exploration than Bootstrapped DQN, from the simple use of an actor with several off-policy critics, illustrates how amenable the architecture of BDPI is to relatively advanced features. We believe that further work will allow more features to naturally emerge, or be easily implemented, on top of the BDPI algorithm we present in this paper.

\section{Robustness to Hyper-Parameters}
\label{sec:app_robustness}

Evaluating the robustness of an algorithm to its hyper-parameters is challenging, and typically not done in Deep RL research. We propose a simple approach, that we designed to be easy to understand and intuitive, and that provides two measures of robustness.

\subsection{Data Collection}

For each algorithm, namely BDPI, DQN, Prioritized and Dueling DQN, A2C, PPO and ACKTR, we define a configuration space that consists of all the combinations of the most relevant hyper-parameters of the algorithms. We then randomly sample configurations, run the algorithm on \textit{LunarLander} for 800 episodes, and compute the total reward obtained during these 800 episodes. We used the OpenAI Baselines implementations of all the algorithms (but BDPI), to ensure that no implementation error on our side invalidates the results.

The hyper-parameters evaluated for each algorithms are listed below. We ensured that all the \textit{known-good} configurations of all the algorithms, for various environments in the literature, are covered.

\begin{description}
	\item[All algorithms] \hfill \\
		\vspace{-5mm}
		\begin{itemize}
			\item[-] Neural network learning rate: 0.00001, 0.00005, 0.0001, 0.0005, 0.001
			\item[-] Neurons in the hidden layer of the neural network: 32, 64, 96, 128, 256
		\end{itemize}
	\item[All but BDPI] \hfill \\
		\vspace{-5mm}
		\begin{itemize}
			\item[-] Number of parallel environments: 1. BDPI is single-threaded, so, to avoid artificially increasing the sensitivity of the other algorithms, we chose to keep this highly-sensitive parameter to 1.
			\item[-] Entropy regularization: 0, 0.01, 0.03, 0.05
		\end{itemize}
	\item[BDPI] \hfill \\
		\vspace{-5mm}
		\begin{itemize}
			\item[-] Experience buffer size: 5K, 10K, 20K, 50K, 100K
			\item[-] Batch size: 64, 128, 256, 512
			\item[-] Critics trained per time-step: 1, 4, 8
			\item[-] Number of critics: 1, 4, 8, 16, 32
			\item[-] Clipped DQN iterations per critic-time-step: 1, 2, 4, 8
			\item[-] Epochs used to fit the neural networks: 1, 4, 8, 16. The absolute best performance of BDPI is achieved with 20-50+ epochs, but our computing resources did not allow us to increase this parameter as much. We ensure that the best-known configuration of the other algorithms is included in our configuration space.
		\end{itemize}
	\item[PPO] \hfill \\
		\vspace{-5mm}
		\begin{itemize}
			\item[-] Steps per batch: 64, 128, 256, 384, 512, 1024, 2048
			\item[-] Lambda: 0.7, 0.8, 0.9, 1.0
			\item[-] Optimization steps per epoch: 1, 2, 4, 8, 16
		\end{itemize}
	\item[A2C] \hfill \\
		\vspace{-5mm}
		\begin{itemize}
			\item[-] Time-steps between learning epochs: 1, 2, 4, 6, 8
			\item[-] Critic loss weight compared to the actor: 0.1, 0.3, 0.5, 0.7, 0.9
			\item[-] Gradient norm clipping: 0.1, 0.3, 0.5, 0.8, 1.0
		\end{itemize}
	\item[ACKTR] \hfill \\
		\vspace{-5mm}
		\begin{itemize}
			\item[-] Learning rate (specific to ACKTR, default of 0.25): 0.01, 0.10, 0.25, 0.50, 0.90
			\item[-] Time-steps between learning epochs: 1, 5, 10, 20, 40, 80
			\item[-] Critic loss weight: 0.1, 0.3, 0.5, 0.7, 0.9
			\item[-] Fisher weight: 0.1, 0.3, 0.5, 0.7, 0.9
			\item[-] Gradient norm clipping: 0.1, 0.3, 0.5, 0.8, 1.0
			\item[-] Kronecker-Factored clipping: 0.0001, 0.001, 0.005, 0.01, 0.1 description
		\end{itemize}
	\item[DQN] \hfill \\
		\vspace{-5mm}
		\begin{itemize}
			\item[-] Experience buffer size: 5K, 10K, 20K, 50K, 100K
			\item[-] Batch size: 16, 32, 128
			\item[-] Exploration fraction: 0.02, 0.05, 0.1, 1.0
			\item[-] Final epsilon after exploration: 0.1, 0.05, 0.01, 0.001
			\item[-] Time-steps between learning epochs: 1, 2, 4, 8, 16
			\item[-] Time-steps before learning starts: 1, 500, 1000, 10000
			\item[-] Target network update frequency: 1, 50, 100, 500, 1000
		\end{itemize}
	\item[Dueling DQN with Prioritized Experience Replay] \hfill \\
		\textit{All the same parameters as DQN, and:}
		\begin{itemize}
			\item[-] Alpha parameter for Prioritized ER: 0.5, 0.6, 0.7, 0.8, 0.9
		\end{itemize}
\end{description}

\subsection{Data Processing}

\begin{figure}
	\centering
	\includegraphics{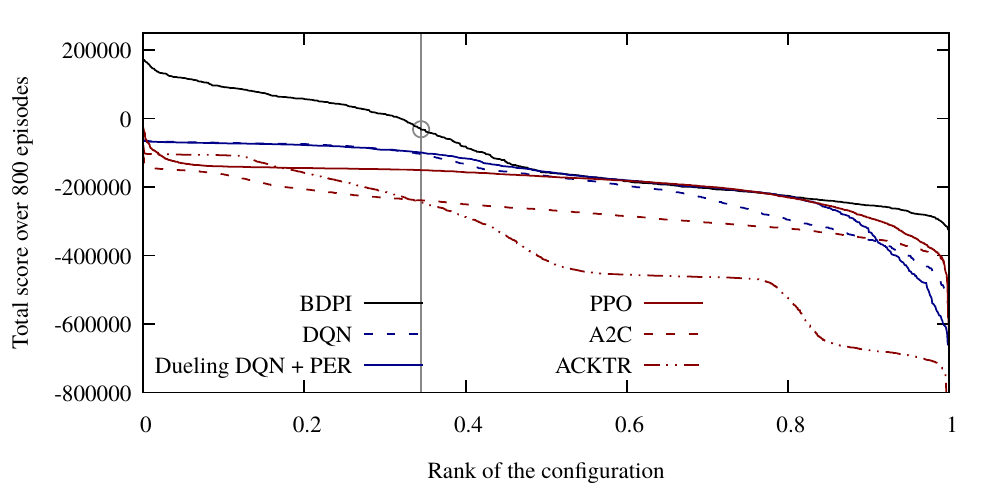}
	\caption{Total reward per configuration, sorted by descending total reward. This plot shows that more than 35\% of BDPI's randomly-sampled configurations perform better than the best (PPO) configuration. The worst BDPI configuration is also better than most of the configurations of the other algorithms. On LunarLander, for 800 episodes, the random policy achieves a total reward of about -240K.}
	\label{fig:robustness}
\end{figure}

Hundreds of randomly-sampled configurations of the algorithms are evaluated, and we propose to use the total reward over 800 episodes on \textit{LunarLander} as performance measure. Figure \ref{fig:robustness} graphically displays this dataset: for each algorithms, all the configurations are sorted by descending total reward, then the lines are stretched horizontally to compensate for the unequal amount of configurations that each algorithm was evaluated on, due to each algorithm having different computational resources requirements.

The measures that we report in Section \ref{sec:exp_robustness} are slightly more advanced. While Figure \ref{fig:robustness} intuitively shows that BDPI produces a higher curve, sorting the configurations by performance remove any information about the locality of the configurations. It shows that many configurations are good, not that they are close together in configuration space. In order to better measure how slight changes in parameters influences performance, be introduce a second measure:

\begin{align}
	\nonumber
	S &= \frac{\sum\limits_{a, b} w(a, b) \delta(a, b)}{\sum\limits_{a, b} w(a, b)} \\
	\label{eq:robustness}
	w(a, b) &= |a_{params} - b_{params}|^{-1} \\
	\nonumber
	\delta(a, b) &= |a_{score} - b_{score}|
\end{align}

\noindent
with $a$ and $b$ two randomly-sampled configurations. In order to produce accurate scores, we evaluate each algorithm on more than 2000 configurations, and apply Equation \ref{eq:robustness} on 4000000 pairs of configurations. The resulting scores, also reported in Section \ref{sec:exp_robustness}, are DQN with Prioritized ER (930), BDPI (1167), vanilla DQN (1326), then, significantly larger, A2C (2369), PPO (2452) and ACKTR (5815).

\section{Experimental Setup}
\label{sec:app_algos}

All the algorithms evaluated in Section \ref{sec:experiments} use feed-forward neural networks to represent their actor(s) and critic(s). They take as input the one-hot encoding of the current state, and are trained with the Adam optimizer \citep{kingma2014adam}, using a learning rate of 0.0001 (0.001 for PPO, as it gave better results). We configured each algorithm following the recommendations in their respective papers, and further tuned some parameters to the environments we use. These parameters are given Table \ref{fig:params}. They are kept as similar as possible across algorithms, and constant across the three sensors-based environments, to evaluate the generality of the algorithms. For \textit{Hallway}, differents sets of parameters have been used (especially for PPO and ACKTR), as explained in Section \ref{sec:exp_algorithms}.

\begin{table}
	\aboverulesep=0pt
	\belowrulesep=0pt
	\renewcommand{\arraystretch}{1.2}
	\begin{tabularx}{\textwidth}{|l|Y|Y|Y|YY|}
		\toprule
		& \rotatebox{90}{ACKTR} & \rotatebox{90}{PPO} & \rotatebox{90}{BDQN} & \rotatebox{90}{ABCDQN~} & \rotatebox{90}{BDPI} \\
		\midrule
		Discount factor $\gamma$ & \multicolumn{5}{c|}{0.99} \\
		\cmidrule{2-6}
		Replay buffer size & -- & -- & 20K & \multicolumn{2}{c|}{20K} \\
		Experiences/batch & 20 & 256/1024$^{(a)}$ & 256 &  \multicolumn{2}{c|}{256} \\
		Training epoch every $K$ time-steps & 20 & 256/1024$^{(a)}$ & 1 &  \multicolumn{2}{c|}{1} \\
		\midrule
		Policy loss & PG+Fisher & PPO & -- & -- & MSE \\
		Trust region $\delta$ & -- & -- & -- & -- & 0.05 \\
		Entropy regularization & 0.01 & 0.01 & -- & -- & 0 \\
		Value loss coefficient & 0.5 & -- & -- & \multicolumn{2}{c|}{--} \\
		\midrule
		Critic count $N_c$ & 1 & 1 & 16 & \multicolumn{2}{c|}{16} \\
		Critic sampling frequency & -- & -- & episode & \multicolumn{2}{c|}{--} \\
		Critic learning rate $\alpha$ & 0.25 & 1.0 (on $R_t$) & 1.0 & \multicolumn{2}{c|}{0.2} \\
		Critic training iterations $N_t$ & -- & 1 & 1 & \multicolumn{2}{c|}{4} \\
		\midrule
		Gradient steps/batch & 1 & 4 & 20 & \multicolumn{2}{c|}{20} \\
		Learning rate & \textit{dynamic} & 0.001 & 0.0001 & \multicolumn{2}{c|}{0.0001} \\
		Activation function & tanh & tanh & tanh & \multicolumn{2}{c|}{tanh} \\
		Hidden layers & 2 & 2 & 1 & \multicolumn{2}{c|}{1} \\
		\cmidrule{2-6}
		Hidden neurons & \multicolumn{5}{c|}{32/256$^{(a)}$} \\
		\bottomrule
	\end{tabularx}
	\caption{Hyper-parameter of the various algorithms we experimentally evaluate. (a) Hyper-parameters that were required for \textit{LunarLander} to perform well.}
	\label{fig:params}
\end{table}

\end{document}